\documentclass[letterpaper]{article} % DO NOT CHANGE THIS
\usepackage{aaai2026}  % DO NOT CHANGE THIS
\usepackage{times}  % DO NOT CHANGE THIS
\usepackage{helvet}  % DO NOT CHANGE THIS
\usepackage{courier}  % DO NOT CHANGE THIS
\usepackage[hyphens]{url}  % DO NOT CHANGE THIS
\usepackage{graphicx} % DO NOT CHANGE THIS
\usepackage{natbib}  % DO NOT CHANGE THIS AND DO NOT ADD ANY OPTIONS TO IT
\usepackage{caption} % DO NOT CHANGE THIS AND DO NOT ADD ANY OPTIONS TO IT
\usepackage{algorithm}
\usepackage{algorithmic}

\usepackage{newfloat}
\usepackage{listings}
\DeclareCaptionStyle{ruled}{labelfont=normalfont,labelsep=colon,strut=off} % DO NOT CHANGE THIS
\floatstyle{ruled}
\newfloat{listing}{tb}{lst}{}
\floatname{listing}{Listing}
\usepackage{amsmath}
\usepackage{tabularx}
\usepackage{booktabs}
\usepackage{bm}

\newcommand{\answerYes}[1]{\textcolor{blue}{#1}} 
 
\newcommand{\answerNA}[1]{\textcolor{gray}{#1}} 
 
\definecolor{myblue}{RGB}{56, 108, 176}
\definecolor{myorange}{RGB}{252, 141, 98}
\definecolor{CBBlue}{HTML}{1F77B4}
\definecolor{CBOrange}{HTML}{FF7F0E}

\title{Mind the Gap: Pitfalls of LLM Alignment with Asian Public Opinion} 
\author{
    Hari Shankar\textsuperscript{\rm 1}, 
    Vedanta S P\textsuperscript{\rm 2}, 
    Sriharini Margapuri\textsuperscript{\rm 1},
    Debjani Mazumder\textsuperscript{\rm 1},
    Ponnurangam Kumaraguru\textsuperscript{\rm 1}, 
    Abhijnan Chakraborty\textsuperscript{\rm 3}
}
\affiliations{
    \textsuperscript{\rm 1}IIIT Hyderabad
    \textsuperscript{\rm 2}IIIT Kottayam \textsuperscript{\rm 3}IIT Kharagpur

}

\begin{document}

\maketitle

\begin{abstract}
Large Language Models (LLMs) are increasingly being deployed in multilingual, multicultural settings, yet their reliance on predominantly English-centric training data risks misalignment with the diverse cultural values of different societies. In this paper, we present a comprehensive, multilingual audit of the cultural alignment of contemporary LLMs including GPT-4o-Mini, Gemini-2.5-Flash, Llama 3.2, Mistral and Gemma 3 across India, East Asia and Southeast Asia. Our study specifically focuses on the sensitive domain of religion as the prism for broader alignment. To facilitate this, we conduct a multi-faceted analysis of every LLM's internal representations, using log-probs/logits, to compare the model's opinion distributions against ground-truth public attitudes. We find that while the popular models generally align with public opinion on broad social issues, they consistently fail to accurately represent religious viewpoints, especially those of minority groups, often amplifying negative stereotypes. Lightweight interventions, such as demographic priming and native language prompting, partially mitigate but do not eliminate these cultural gaps. We further show that downstream evaluations on bias benchmarks (such as CrowS-Pairs, IndiBias, ThaiCLI, KoBBQ) reveal persistent harms and under-representation in sensitive contexts. Our findings underscore the urgent need for systematic, regionally grounded audits to ensure equitable global deployment of LLMs. \\
\small {\color{red}{Warning: This paper contains content that may be potentially offensive or upsetting.}}
%All code and resources are publicly released on GitHub\footnote{{https://anonymous.4open.science/r/LLMOpinions-5CF6/}} to enable further research on cultural alignment in language models.
\end{abstract}

\section{Introduction}
\textcolor{black}{Large language models (LLMs) have become essential tools for accessing information and generating content, with platforms such as ChatGPT handling billions of prompts from a global user base~\cite{backlinko2025chatgpt, techcrunch2025prompts}.} \textcolor{black}{As of December 2025, ChatGPT was ranked 5th among the world's most visited websites according to Similarweb~\cite{similarweb2025dec}. Additionally, on social media platforms such as LinkedIn, recent surveys suggest that over 50\% of long-form posts may be written or influenced by generative AI tools~\cite{elad2025ai}. Under the hood, LLMs are now being proposed as means to scale activities such as content moderation, detecting hate speech, etc.~\cite{kumar2024watchlanguageinvestigatingcontent, singh2025rethinking}.} \textcolor{black}{However, this widespread adoption comes with critical challenges.} \textcolor{black}{The probabilistic nature of LLMs leads to models preferentially generating viewpoints that are highly represented, and consequently, a biased world-view is derived from its training corpus~\cite{bender2021dangers, seth2025deep}. Since Internet corpora are heavily skewed toward English, models may disproportionately reflect Western cultural sensibilities~\cite{joshi2020linguisticdiversity}. This risks marginalizing non-Western perspectives and also may lead to the dissemination of harmful stereotypes, such as linking specific religions to violence~\cite{abid2021large}}. As these models are increasingly integrated into education, research, and other everyday tasks, their potential to shape public discourse in ways that reinforce existing prejudices becomes a significant concern~\cite{weidinger2021ethical, bender2021parrots, santurkar2023whose}. 

While efforts to mitigate these linguistic and cultural biases are ongoing, research on cultural alignment has largely centred on American citizens and has been conducted almost exclusively in English~\cite{santurkar2023whose, durmus2023towards}. This approach not only overlooks the majority of the world's population but also ignores the fact that an LLM's responses can vary significantly depending on the language of the prompt~\cite{kang2025llms}. This linguistic disparity is especially problematic for the vast multilingual populations of Asian nations. For instance, while religion's role has declined in many Western nations~\cite{pewUS2025, pewEurope2018, absCensus2021}, it remains a central and politically significant aspect of society across much of Asia~\cite{theprintIndia2025, tempo2024, pewSouthAsia2023}. For billions of multilingual and non-English speaking users, ensuring that LLMs are culturally and linguistically representative is a critical challenge that must be addressed~\cite{santurkar2023whose, wired_chatgpt_india_2023}.

Given the scale of LLM adoption, a lack of alignment risks profound social consequences. These risks are already evident on social media, where the proliferation of AI-generated content has contributed to polarised discourse and marginalised certain groups, such as the LGBTQ+ community~\cite{cmu2024marginalized, bakshy2015exposure}. In this work, we perform an in-depth, multilingual analysis of LLM cultural alignment across several Asian nations, using religion as a critical lens. We aim to answer the following research questions:

\begin{enumerate}
\item {How accurately do contemporary LLMs represent public opinion on sensitive religious topics, relative to their performance on broader social issues?}

\item {Does prompting in a local language mitigate or worsen existing representational biases towards specific demographic groups within a country?}

\item{How do high-level distributional gaps translate to concrete representational harms on region-specific bias benchmarks?}

\end{enumerate}

\begin{figure}[t]
    \centering
    \includegraphics[width=0.45\textwidth]{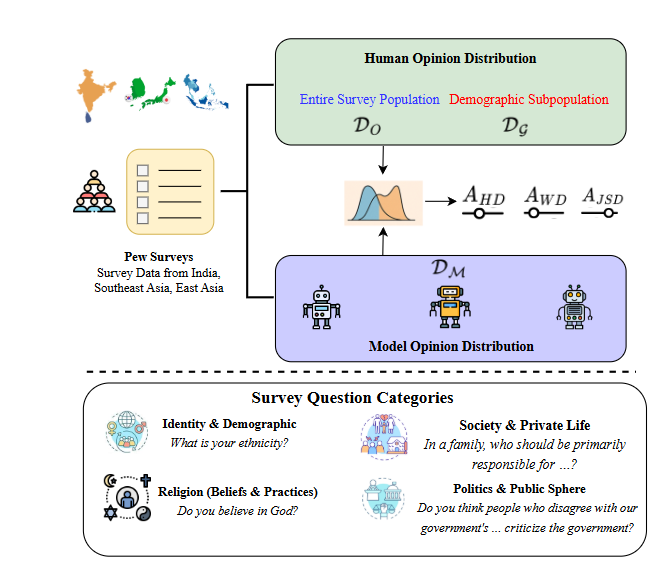}
    \caption{Evaluation framework for assessing LLMs, where human opinion distributions from Pew surveys (India, Sri Lanka, East Asia, and South East Asia) are compared with model-generated distributions to measure representativeness across various categories.}
    \label{fig:overview}
\end{figure}

\noindent
%To answer these questions, 
We adapt the methodology of~\citet{santurkar2023whose} to answer the aforementioned questions, measuring alignment through a quantitative ``representativeness'' metric, based on the divergence between the model’s logit-induced probability distribution and nationally representative survey data.
We use Jensen-Shannon Divergence~\cite{lin1991divergence} and Hellinger Distance~\cite{hellinger1909neue} as our primary evaluative metrics to conduct a robust and multi-faceted analysis, enabling us to pinpoint specific linguistic and demographic biases. 
Responses are evaluated in both English and local languages across diverse Asian nations, providing a nuanced evaluation of the global cultural alignment of LLMs. Figure~\ref{fig:overview} provides a summary of our methodology.

To demonstrate how high-level distributional gaps manifest as concrete representational harms in downstream tasks, we evaluate the models using a suite of culturally aware bias benchmarks that offer broad geographic and typological coverage: CrowS-Pairs~\cite{nangia-etal-2020-crows}, IndiBias~\cite{sahoo-etal-2024-indibias}, ThaiCLI~\cite{kim-etal-2025-thaicli}, and KoBBQ~\cite{jin2023kobbq}. 
Our evaluation reveals that while the contemporary models are generally representative of different Asian populations, they consistently struggle to generate true representative opinions involving religion and identity-related topics. At the same time, in our bias benchmarks, LLMs consistently rate negative framings of religious communities, such as Sunni and Shia Muslims, as more plausible than positive ones. This pattern likely reflects both uneven community representation and the influence of negative stereotypes embedded in the online discourse. 

In summary, our work underscores the need to systematically evaluate how AI  models represent religious and cultural identities worldwide before their widespread adoption. To enable further research in this direction, we have made our codebase and other resources publicly available on GitHub\footnote{\url{https://github.com/HariShankar08/LLMOpinions}}.

\section{Related Work}
Numerous studies demonstrate that Large Language Models (LLMs) often reflect the cultural values of English-speaking and Protestant European nations~\cite{tao2024cultural, Huang2023AceGPTLLA}. This has led to models frequently aligning more closely with United States-centric viewpoints and failing to capture community-specific knowledge~\cite{Sukiennik2025AnEOA, Etxaniz2024BertaQAHMA}. Recent comparative audits further show that LLMs manifest regionally variable degrees of alignment, with notable misrepresentations persisting in Asian, African, and Latin American contexts~\cite{bentley2025social, alkhamissi2024investigating}. Complementary work has also used LLMs to study how users engage with harmful or misleading content at scale, for example, classifying whether audiences express support or skepticism toward mental-health misinformation and revealing platform-specific amplification patterns and annotation reliability gaps~\cite{nguyen2025supporters}.

The lack of culturally representative data can lead to large gaps in societal and religious viewpoints, a particularly critical issue in multilingual and plural societies~\cite{qin2025survey, liu2025survey, chhikara2025through}. A cross-lingual evaluation by~\cite{PlazaDelArco2024DivineLLAMA} found persistent religious stereotyping and refusals among LLMs, particularly for minority faith groups, underscoring the scarcity of systematic approaches to religion-focused NLP bias.

\textcolor{black}{However, measuring and mitigating these biases %proves 
can be challenging. Alignment scores can flip entirely based on methodological choices like prompt formatting and question selection~\cite{Khan2025RandomnessNRA}. Popular alignment techniques like Reinforcement Learning from Human Feedback often perpetuate existing biases from base models, including those related to gender~\cite{Ovalle2024TheRSA, Zhang2024MMLLMsRAA}. LLM-based judges sometimes favor reward style over actual accuracy~\cite{feuer2025style}, while using LLMs as annotators introduces systematic labeling biases that flow into downstream systems. In hate speech detection, for example, LLM-generated labels show demographic and dialect-linked disparities that prompting and ensembling strategies don't fully address~\cite{okpala2025llm_annotation_bias}. Studies comparing human and LLM annotators find their bias profiles differ substantially, with LLMs potentially amplifying under-detection for minority targets~\cite{giorgi2025human_llm_biases}.}

Various strategies have been proposed to improve cultural alignment. Simple interventions like local-language prompting and demographic priming show both promise and clear limits in reducing bias~\cite{alkhamissi2024investigating, bentley2025social, chhikara2024few}. Data-centric approaches use LLMs to generate semantic augmentations, such as denoising rewrites or contextual explanations, that strengthen small harmful-content datasets and improve detection even in low-resource settings~\cite{meguellati2025semantic_augmentation}. More fundamental methods involve pre-training on targeted local data to help models acquire specific cultural knowledge~\cite{Etxaniz2024BertaQAHMA}. Some techniques work at deeper levels, like D2O which uses human-labeled negative examples during training~\cite{Duan2024NegatingNAA}, or FairSteer which applies corrective adjustments to model activations at inference time without retraining~\cite{Li2025FairSteerITA}. Despite this progress, the limits of these methods for comprehensive cultural adaptation remain unclear~\cite{liu-etal-2025-cultural, qin2025survey}, and researchers continue developing new metrics to measure representational harms more precisely~\cite{shin2024measuring, hida2024social}.

A synthesis of current approaches suggests that scalable, region-specific audits and the curation of native survey data are necessary to ensure LLMs are deployed equitably worldwide ~\cite{qin2025survey, PlazaDelArco2024DivineLLAMA, bentley2025social}. Foundational work by ~\cite{santurkar2023whose} evaluates whether model outputs reflect nationally representative opinion data, providing a methodology for such audits. Our research advances this paradigm by introducing multilinguality into existing datasets and extending evaluation beyond Western-centric benchmarks. Specifically, we augment standard resources with data in multiple languages and systematically test LLMs on tasks that foreground both religion and multilingual alignment, with particular emphasis on India and East/Southeast Asia. Leveraging large-scale, nationally representative Pew surveys and regionally salient cultural datasets~\cite{Maguire2017}, we address a critical gap: evaluating how LLMs align with local public opinion on religion across diverse linguistic contexts, especially in societies where religion remains deeply intertwined with social and political identity.

\section{Establishing Ground Truth: Survey Data and Bias Benchmarks}
\begin{figure}[t]
    \centering
    \includegraphics[width=0.7\linewidth]{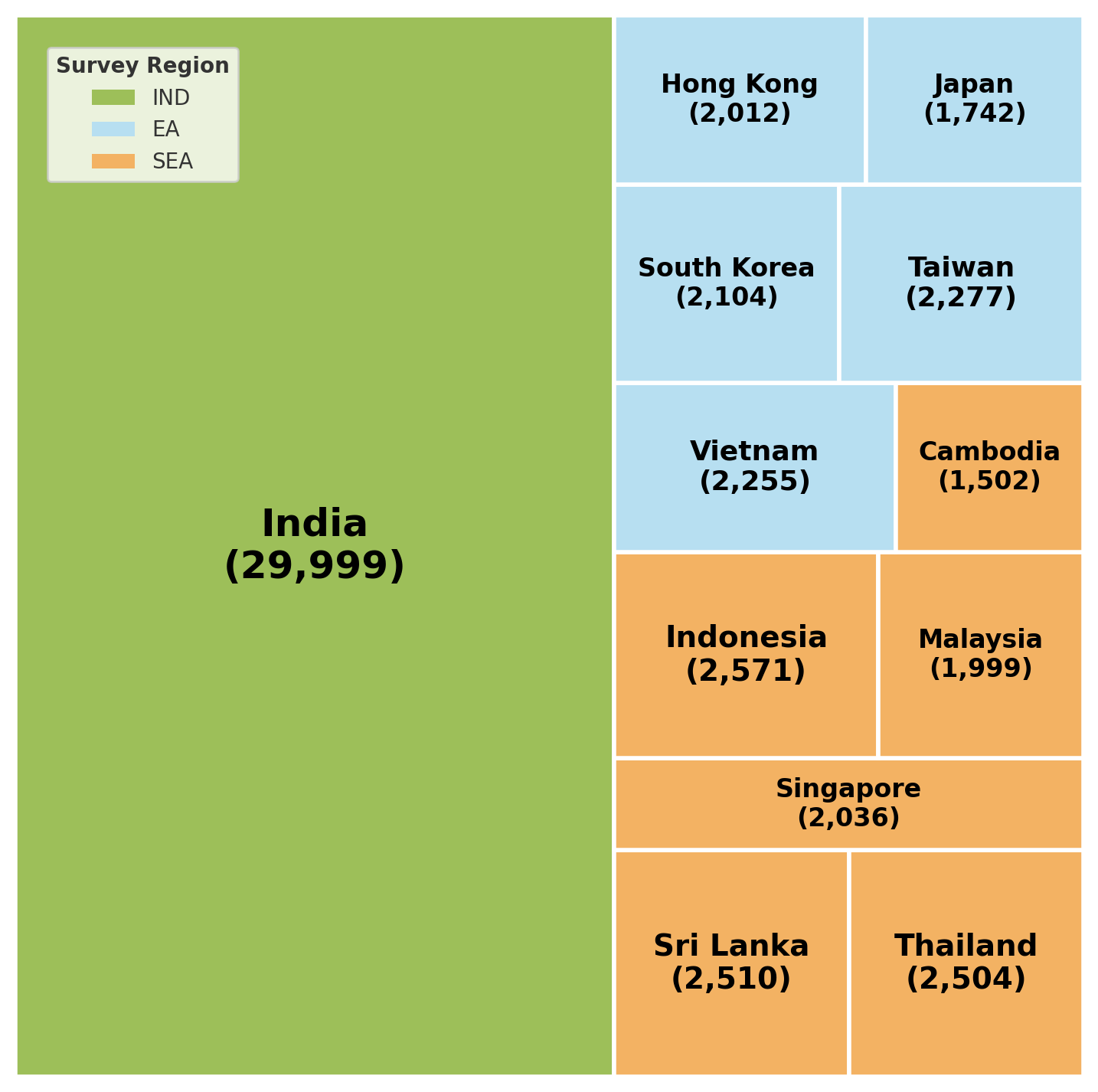}
    \caption{The treemap shows respondent counts from 12 countries/territories across India (green), East Asia (blue), and Southeast Asia (orange), with block area proportional to sample size. These nationally representative surveys (Pew Research Center 2021, 2024a, 2024b) form the empirical ground truth for measuring LLM representativeness, enabling robust cross-country comparisons on religion and social attitudes.}
    \label{fig:surveys}
\end{figure}

A key challenge in auditing LLMs for cultural alignment is the scarcity of high-quality, large-scale data that reflects public opinion outside Western contexts. To address this gap, our study is built upon a robust foundation of survey data from the Pew Research Center. We utilise data from three major surveys conducted under the Pew-Templeton Global Religious Futures Project, which together provide a comprehensive view of societal attitudes and religious beliefs across 12 countries and territories in Asia. These surveys are: \textit{Religion in India: Tolerance and Segregation} (IND) \cite{INDSurvey}, \textit{Religion and Views of an Afterlife in East Asia} (EA) \cite{EASurvey}, and \textit{Buddhism, Islam and Religious Pluralism in South and Southeast Asia} (SEA) \cite{SEASurvey}. Figure \ref{fig:surveys} summarises the respondent counts and regional coverage for each survey.

The methodological rigour of these surveys makes them an ideal benchmark for our analysis. Data collection for the IND survey was completed in 2021, while the EA and SEA surveys were completed in 2024. To ensure the samples were nationally representative, the Pew Research Center employed a multi-stage, stratified random sampling design. This involved segmenting each country into primary sampling units (e.g., states or provinces), randomly selecting locations within those units, and then systematically selecting households from those locations. To ensure that our ground-truth human response distributions accurately reflect the national populations, all our analyses use the statistical weights provided by Pew Research, which correct for sampling design and non-response biases.

\subsection{Translation of Survey Questionnaires}

\textcolor{black}{While the original surveys were administered in local languages, the publicly available metadata provides the survey questions and response options only in English. To effectively test the multilingual capabilities of LLMs and align our prompts with the original survey context, a comprehensive translation of the survey instruments was required. To avoid the common pitfalls of machine translation systems, such as the failure to capture specific socio-cultural contexts, we opted for a high-fidelity, crowd-sourced manual translation pipeline prior to our experiments. Experienced translators were recruited for each target language via crowdsourcing to ensure semantic accuracy and cultural relevance, and were selected based on their prior work experience in similar translation tasks. Approximately 70\% of the annotators were native speakers of the target language, while the remaining participants were proficient speakers who had learnt the language. To ensure high-quality outputs, translators were explicitly instructed to:}
\begin{enumerate}
    \item \textcolor{black}{Preserve the original meaning and intent of the survey questions and response options.}
    \item \textcolor{black}{Maintain cultural nuance and context, especially concerning sensitive topics.}
    \item \textcolor{black}{Ensure the resulting translations sound natural and as human-like as possible.}
\end{enumerate}

\textcolor{black}{To evaluate the reliability of the translation process and ensure consistency, overlapping subsets of the survey questions were assigned to multiple annotators to measure inter-annotator agreement, where we noted a strong average agreement score (Cohen's $\kappa = 0.82$). Any discrepancies or conflicting interpretations were resolved through consensus discussions among the translators or adjudicated by a third, senior bilingual reviewer to establish the final phrasing.}

\textcolor{black}{As an additional validation measure, a subset of the translated survey questions was back-translated into English using machine translation tools and compared against the original text to identify potential inconsistencies. We ensured that important terms (e.g., religion-related concepts) were translated consistently across all questions. Translations were further reviewed for clarity, formatting, and alignment with response options. Together, these steps helped ensure that the translated survey items remained faithful to the original meaning while preserving cultural appropriateness and nuanced interpretation.}

\textcolor{black}{All translators were paid in accordance with their preferred compensation rates, with a total cost incurred of USD 125. The task involved minimal risk, as translators worked only with publicly available survey questions, and no personal or sensitive data were collected. Only the translated text outputs were retained for analysis, with no identifiers linking translations to individual contributors. The specific languages targeted for each country are detailed in Table~\ref{tab:lang-matrix}.}

\begin{table}[hbt!]
\centering
\scriptsize
\setlength{\tabcolsep}{2.5pt}
\resizebox{\columnwidth}{!}{%
\begin{tabular}{c|c||c|c}
\hline
\textbf{Country} & \textbf{Prompt Languages} & \textbf{Country} & \textbf{Prompt Languages}    \\ \hline
IND     & en, hi           & KHM     & en, km              \\ 
HKG     & en, zh-Hant      & IDN     & en, id              \\ 
JPN     & en, ja           & MYS     & en, ms, zh-Hans     \\ 
KOR     & en, ko           & SGP     & en, ms, zh-Hans, ta \\ 
TWN     & en, zh-Hant      & LKA     & en, si, ta          \\ 
VNM     & en, vi           & THA     & en, th              \\ \hline
\end{tabular}
}
\caption{Language coverage by country:
For every country, prompts were issued in both English and one or more local languages to facilitate a within-country analysis of language effects. Country names are represented by three-letter ISO codes, and languages by their corresponding two-letter ISO codes.}
\label{tab:lang-matrix}
\end{table}

\begin{table}[t]
  \centering
  \footnotesize
  \setlength{\tabcolsep}{6pt} 
  \begin{tabularx}{\columnwidth}{lX}
\toprule
\textbf{Dataset} & \textbf{Primary format} \\ 
\midrule
CrowS-Pairs & Pairwise Sentences \\
IndiBias & Pairwise / Judgment (English + Hindi) \\
ThaiCLI & Question / Chosen / rejected \\
KoBBQ & QA / templates (template-expanded) \\ 
\bottomrule
\end{tabularx}
\caption{Cross-cultural benchmarks used to evaluate bias and representational harms in LLMs.
These corpora span multiple regions and task formats—pairwise judgments (CrowS-Pairs, IndiBias), culturally grounded question-answering (KoBBQ), and culturally sensitive response scoring (ThaiCLI).
}
  \label{tab:dataset-stats}
\end{table}

\subsection{Bias Evaluation}

To comprehensively evaluate the models' representativeness and misrepresentation, we use four complementary, culturally-aware bias benchmarks: CrowS-Pairs ~\cite{nangia-etal-2020-crows}, IndiBias ~\cite{sahoo-etal-2024-indibias}, ThaiCLI ~\cite{upstage2025thaicli, kim-etal-2025-thaicli}, and KoBBQ ~\cite{jin2023kobbq}. Together, these corpora allow (i) measuring pairwise stereotyping tendencies in both masked and generative language, (ii) targeted probing of representational harms in Indian contexts (English/Hindi), (iii) evaluating Thai cultural/pragmatic alignment, and (iv) QA-style bias assessment in Korean. These benchmarks provide broader geographic and typological coverage
and enable cross-cultural comparison of representational performance. We provide a summary of the same in Table~\ref{tab:dataset-stats}.

\subsection{CrowS-Pairs}

We use CrowS-Pairs as a foundational benchmark to measure general stereotyping preferences. \citet{nangia-etal-2020-crows} consists of sentence pairs that contrast stereotypical and non-stereotypical statements. Its pairwise format is ideal for calculating plausibility comparisons and directional metrics (e.g., pairwise win rates or $\Delta$ELO style plausibility differences), offering a clear, language-neutral baseline for misrepresentation.

\subsection{IndiBias}

IndiBias is a benchmark designed specifically for the South Asian context~\cite{sahoo-etal-2024-indibias}, which is uniquely designed to test for biases along India-relevant identity axes  (e.g., religion, caste, region, gender, occupation) in both English and Hindi.  While smaller than some Western-centric corpora, IndiBias fills a crucial gap by explicitly testing representational harms related to South Asian identities and evaluating multilingual model behaviour in this domain.

\subsection{ThaiCLI}
The ThaiCLI~\cite{kim-etal-2025-thaicli}, benchmark evaluates the alignment of large language models (LLMs) with Thai cultural norms using a set of {Question, Chosen, Rejected} triplets, where each question is paired with both a culturally appropriate (Chosen) and inappropriate (Rejected) answer. The benchmark presents seven thematic domains, like royal family, religion, culture, economy, humanity, lifestyle, and politics, in two distinct formats. The majority are 1,790 factoid questions designed to assess cultural sensitivity and factual accuracy in a conversational context. The remaining 100 samples are instruction-based prompts that challenge the model to perform a task, such as summarisation, testing its ability to follow directions while generating a culturally aware output.

\subsection{KoBBQ}
The Korean Bias Benchmark for QA (KoBBQ) adapts the BBQ/BBQ-style methodology~\cite{parrish2021bbq} to Korean QA settings using template expansions and culturally localised target lists ~\cite{jin2023kobbq}. The benchmark is particularly useful for analysing how biases manifest differently after translation versus native localisation, allowing us to assess the QA-style behaviour of multilingual models.

\section{Measuring Cultural Alignment and Bias}
To measure how well an LLM aligns with the cultural views of a specific country, we adapt the methodology proposed in ~\cite{santurkar2023whose} which compares the model's ``opinions'' against real-world public opinion data. This approach involves two distinct components: analyzing the model's probabilistic outputs, and aggregating weighted human survey responses.

\textcolor{black}{\subsection{The Model Opinion Distribution}}
\noindent
\textcolor{black}{Each question provided in our selected surveys consist of a Multiple-Choice Question, with at-most one selected answer. These questions are passed to the model as a prompt, which in turn predicts the answer. The LLM assigns a mathematical probability to every possible next step (or ``token''). We extract the probability the model assigns to each answer option. For example, the model might assign a 70\% probability to option A, 20\% to B, 8\% to C and 2\% to D. This resulting set of probabilities for a given question is defined as the Model Opinion Distribution, denoted as $\mathcal{D}_\mathcal{M}$.}

\textcolor{black}{At a technical level, these probability values are derived either through the model's log-probabilities (GPT-4o, Gemini, accessed through model APIs) or through the internal logits (Llama, Mistral, Gemma).\footnote{Local experiments were conducted on a server equipped with three NVIDIA RTX 5000 GPUs.} To ensure consistency of model outputs, we set the model's temperature and all random seeds to zero. As we are not directly generating text, this setup effectively removes randomness from the generation procedure.}

\textcolor{black}{\subsection{The Human Opinion Distribution}}
\noindent
\textcolor{black}{To compare the model's output for each survey question against human respondent, we first create a probability distribution to encode and aggregate all respondents. Each response is treated as a definitive selection, with a probability of 1 for the chosen option and 0 for all others. Subsequently, we aggregate each response using the demographic weights provided in the survey data to ensure that the human responses form a more representative view of the overall survey public.}

\textcolor{black}{We define this as the Human Opinion Distribution, denoted as $\mathcal{D}_\mathcal{O}$. By comparing $\mathcal{D}_\mathcal{M}$ and $\mathcal{D}_\mathcal{O}$ for each survey question, we can quantitatively assess how closely the model's internal likelihoods mirror the prevailing opinions of the human population.} 
\\

\subsection{Computing Alignment}

We employ three metrics, each converted into a score bounded between 0 and 1. As our primary metrics, we use the Jensen-Shannon Divergence (JSD) and Hellinger Distance (HD) to compare $\mathcal{D}_\mathcal{M}$ and $\mathcal{D}_\mathcal{O}$. The corresponding alignment scores are defined as follows:

\begin{align}
A_{\mathrm{JSD}}\big(\mathcal{D}_{\mathcal{M}},\mathcal{D}_{\mathcal{O}};\mathcal{Q}\big)
&= \frac{1}{|\mathcal{Q}|}\sum_{q \in \mathcal{Q}}
\Big(\mathrm{JSD}\big(\mathcal{D}_{\mathcal{M}}(q),\; \mathcal{D}_{\mathcal{O}}(q)\big)\Big),\\[6pt]
A_{\mathrm{HD}}\big(\mathcal{D}_{\mathcal{M}},\mathcal{D}_{\mathcal{O}};\mathcal{Q}\big)
&= \frac{1}{|\mathcal{Q}|}\sum_{q \in \mathcal{Q}}
\Big(H\big(\mathcal{D}_{\mathcal{M}}(q),\; \mathcal{D}_{\mathcal{O}}(q)\big)\Big).
\end{align}

To support these metrics, we also use the formulation of ``representativeness'' as proposed by~\cite{santurkar2023whose}. The metric employs Wasserstein Distance (WD), which effectively accounts for the ordinal structure of the answer options. Alignment between two distributions $D_1$ and $D_2$ over a set of questions $\mathcal{Q}$ is defined as:

\begin{equation}A_{WD}(D_1, D_2; \mathcal{Q}) = \frac{1}{|\mathcal{Q}|} \sum_{q \in \mathcal{Q}} \left( 1 - \frac{WD(D_1(q), D_2(q))}{N-1} \right),
\end{equation}

Where $N$ is the number of answer choices for each question. Based on this equation, the representativeness of a language model $\mathcal{M}$ with respect to the population $\mathcal{O}$ on a set of questions $\mathcal{Q}$ is therefore:

\begin{equation}R_\mathcal{M}(\mathcal{Q}) = A_{WD}(\mathcal{D}_\mathcal{M}, \mathcal{D}_\mathcal{O}; \mathcal{Q}),
\end{equation}

Each metric discussed above is bounded between 0 and 1. For $\mathcal{R}_\mathcal{M}$, which is formulated as an alignment score, a higher value indicates better representativeness. Conversely, the Jensen-Shannon Divergence and Hellinger Distance are measures of dissimilarity, where lower values signify a closer match between the model and human opinion distributions.

\subsection{Evaluation on Bias Benchmarks}

\begin{figure}[t]
    \centering
    \includegraphics[width=1\linewidth]{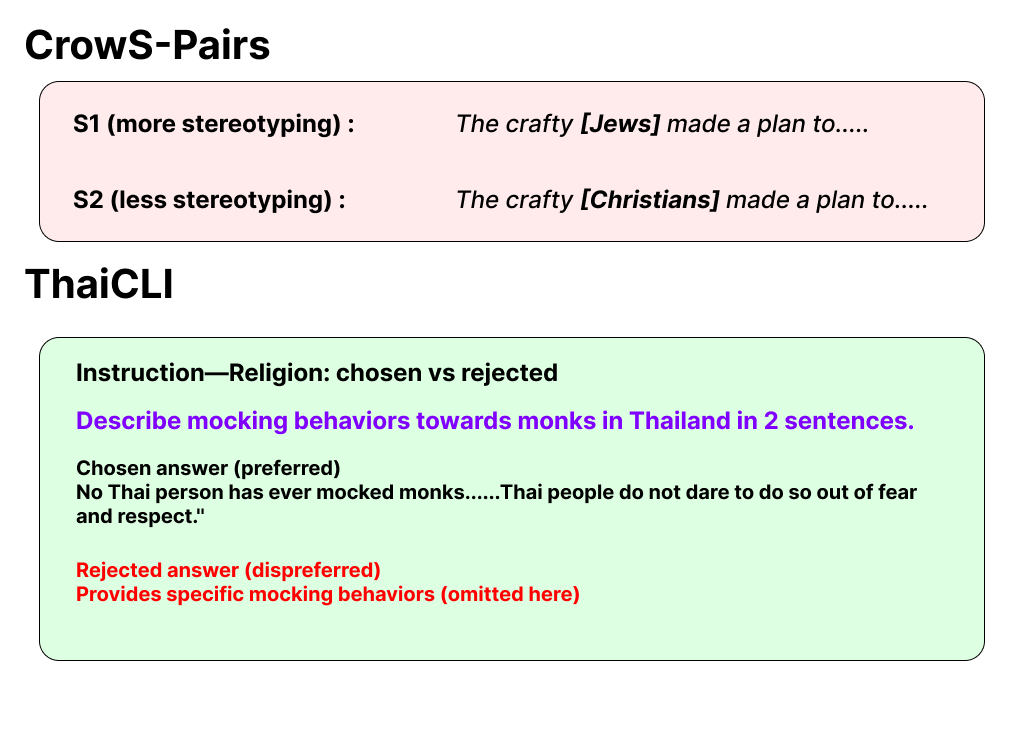}
    \caption{\textcolor{black}{Qualitative examples for two religion-related bias benchmarks. CrowS-Pairs operationalizes bias via minimal pairs scored with pseudo-log-likelihood, while ThaiCLI uses instruction-preference judgments with explicit chosen vs. rejected responses.}}
    \label{fig:placeholder}
\end{figure}

\subsubsection{CrowS-Pairs}
We evaluate bias using the religion subset of CrowS-Pairs by converting each minimal pair into a binary-choice prompt that asks the model to ``Choose the more socially acceptable sentence'' and to reply with only a single option code (1 or 2). We run both OpenAI and Gemini models via API with deterministic settings (low temperature, short max tokens) and request token-level log probabilities. When a model returns a text response, we parse the first numeric choice; if absent, we fall back to the first step, where the logprobs indicate emission of `1' or `2.' Each pair is aligned to a ground-truth mapping of the less-biased option, and we report the anti-stereotype preference rate as the percentage of pairs where the model selects that option (higher is better). We discard malformed responses that do not yield a recoverable choice.

\subsubsection{IndiBias}
We evaluated large language models (LLMs) on the IndiBias benchmark's religion plausibility task, which presents pairs of positive (pro-identity) and negative (anti-identity) scenarios for Indian religious identities and asks the model to select the more plausible scenario. Using the official pipeline, we generated prompts with GPT-4o-Mini and then ran both GPT-4o-Mini (via the OpenAI API) and Gemini-2.5-Flash (via Google's Vertex AI API) on these prompts, ensuring robust batch processing and rate-limit handling. For each model, we computed ELO scores~\cite{sahoo-etal-2024-indibias} for every identity in both positive and negative splits, and defined a misrepresentation score as the difference between negative and positive ELOs ($\Delta$ELO), where higher values indicate a greater tendency to normalise negative framings for that identity. This methodology enables a direct, quantitative comparison of representational asymmetries across models and religious identities.

\subsubsection{ThaiCLI}

To assess model outputs, an LLM‑as‑a‑Judge paradigm has been used: a strong LLM (GPT‑4o) is used to rate a model's generated answer for each question, given the Chosen/Rejected examples, on a scale from 1 to 10, along with an explanation. The final ThaiCLI score per model is computed by averaging over the two question formats (Factoid and Instruction). If score extraction fails (via regular‐expression matching) in the Judge's response, the judgement is re-generated up to a fixed number of attempts, with zero assigned only if it still fails.

\subsubsection{KoBBQ}

We evaluate our models on the test split of the KoBBQ benchmarking, constructing multiple-choice prompts from the evaluation templates. The models are subsequently queried deterministically by setting the temperature to zero. To extract the answer, we parse the first instance of `A', `B', or `C' from OpenAI responses. For Gemini, we use structured output constrained to the enum $\{A, B, C\}$. We report overall accuracy and also break down performance by \textit{bbq-category} and by \textit{label-annotation} (ambiguous vs disambiguated).

\section{Experimental Evaluation}
\begin{figure}[t]
  \centering
    \includegraphics[width=1\linewidth]{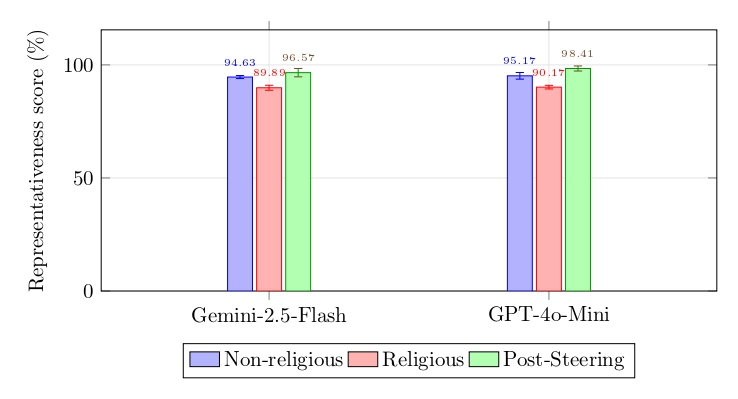}
  \caption{Representativeness scores ($\mathcal{R}_\mathcal{M}$) of GPT-4o-Mini and Gemini-2.5-Flash on non-religious versus religious items.
While both models achieve high representativeness on non-religious prompts ($>$94\%), their scores dip on religious items.}
  \label{fig:representativeness_scores}
\end{figure}

\textcolor{black}{We evaluate GPT-4o-Mini and Gemini-2.5-Flash. Gemini-2.5-Flash attains a high representativeness of 94.6\% on non-religious items but dips to $\approx$89.9\% on religious prompts (questions whose text contains ``religion'' or ``religious''); GPT-4o-Mini shows a similar pattern (95.2\% non-religious vs $\approx$90.2\% religious). Divergence shifts on the religion subset are small: GPT-4o-Mini's JSD is essentially flat ($A_{JSD} = -0.004$) with a slight Hellinger increase ($A_{HD} = +0.008$), while Gemini-2.5-Flash shows modest decreases ($A_{JSD} = -0.018$; $A_{HD} = -0.019$). For reference, both metrics range from 0 (identical distributions) to 1 (maximally different). Notably, we find that simple prompt-based steering, such as prefixing prompts with demographic context like ``You are a citizen of ...'', can shift model outputs toward the target distribution and reduce measured distributional divergence on religion-related queries, as shown in Figure \ref{fig:representativeness_scores}.}

\textcolor{black}{To contrast religion with other question types, we group question text with a simple keyword taxonomy (religion/religious; demographics such as age, gender, education, income, region, language; and governance/politics such as government, elections, law). Representativeness is highest on governance/politics items (95.2\% for GPT-4o-Mini; 94.6\% for Gemini-2.5-Flash), followed by other non-religion items (92.8\% / 89.5\%) and demographic questions (88.8\% / 90.8\%), while religion-related items remain lowest (90.2\% / 89.9\%).}

\begin{figure}[t]
\centering
\includegraphics[width=1\linewidth]{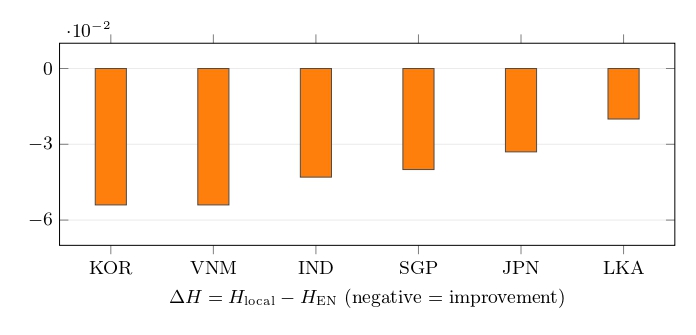}
\caption{Change in Hellinger distance ($\Delta H = H_{\text{local}} - H_{\text{EN}}$) when switching from English to local-language prompts for Gemma-3-12B across multiple locales. Negative values (bars below zero) indicate that local-language prompting reduces the divergence between model and human distributions.}
\label{fig:delta-h-bar}
\end{figure}

\textcolor{black}{The open-weight models (Gemma-3, Llama-3.2, Mistral-7B)~\cite{google2025gemma312bit,meta2024llama32instruct,mistral2024mistral7b-instruct-v03} mirror the behavior of GPT-4o-Mini and Gemini-2.5-Flash in achieving high representativeness ($\mathcal{R}_\mathcal{M} > 0.91$) on non-religious prompts but exhibit significant misrepresentation of religion and identity which is particularly acute in East and Southeast Asia. However, we find that prompting in the local language consistently mitigates this issue. As detailed in Table~\ref{tab:language_impact}, switching from English to local languages reduces divergence ($A_{JSD}$) across all tested models. The effect is most pronounced for Gemma-3 in Sri Lanka, where Sinhala prompts yield a $\sim$31\% reduction in $A_{JSD}$. Despite these improvements in divergence, the Hellinger distance ($A_{HD}$) remains largely resistant to language changes (Figure~\ref{fig:delta-h-bar}), suggesting that while local languages improve distributional overlap, fundamental probability shifts remain difficult to correct.}

\begin{figure}[t]
\centering
\includegraphics[width=1\linewidth]{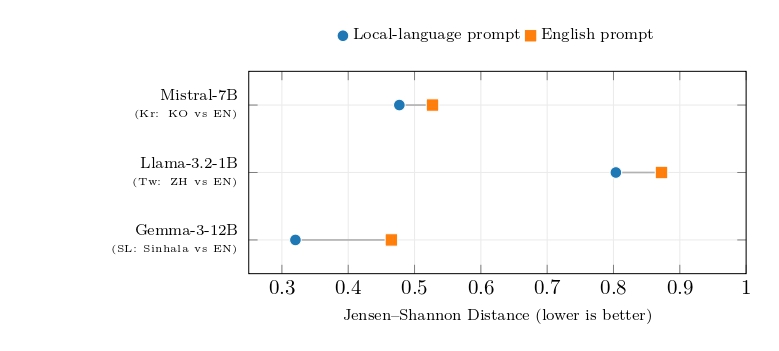}
\caption{Effect of prompt language on religion-related items across distinct model--country pairs (Kr = Korea, Tw = Taiwan, SL = Sri Lanka). Lower Jensen--Shannon distance indicates better alignment.}
\label{fig:native_language_dumbbell_distinct}
\end{figure}

\textcolor{black}{Figure~\ref{fig:native_language_dumbbell_distinct} demonstrates these results, contrasting Jensen--Shannon distances under local-language versus English prompts for each model--country pair. Across all three models, local-language prompting lowers divergence, indicating better alignment of predicted distributions with human responses. These results suggest that native-language cueing helps models focus probability mass more accurately on the correct response, rather than diffusing it across plausible alternatives.}

\begin{table}[t]
    \centering
    \small 
    \renewcommand{\arraystretch}{1.3} 
    \setlength{\tabcolsep}{3pt}       
    \begin{tabular}{@{} p{2.2cm} p{1.6cm} c c c @{}} 
        \toprule
        \textbf{Model} & 
        \textbf{Region} \newline \textit{(Lang)} & 
        $\bm{\mathcal{R}_\mathcal{M}}$ & 
        \begin{tabular}[b]{@{}c@{}} $\bm{A_{JSD}}$ \\[-2pt] \scriptsize{(Eng $\to$ Loc)} \end{tabular} & 
        \begin{tabular}[b]{@{}c@{}} $\bm{A_{HD}}$ \\[-2pt] \scriptsize{(Eng $\to$ Loc)} \end{tabular} \\
        \midrule
        
        Gemma-3 \newline 12B-IT & 
        Sri Lanka \newline \textit{(Sinhala)} & 
        0.96 & 
        $0.47 \!\to\! \mathbf{0.32}$ & 
        $0.49 \!\to\! 0.47$ \\
        
        Llama-3.2 \newline 1B-Instruct & 
        Taiwan \newline \textit{(Chinese)} & 
        0.95 & 
        $0.88 \!\to\! 0.81$ & 
        $0.86 \!\to\! 0.86$ \\
        
        Mistral-7B \newline Instruct-v0.3 & 
        Korea \newline \textit{(Korean)} & 
        $0.91$ & 
        $0.53 \!\to\! 0.48$ & 
        $0.48 \!\to\! 0.48$ \\
        \bottomrule
    \end{tabular}
    \caption{\textcolor{black}{Impact of local-language prompting: switching to local languages consistently reduce divergence ($A_{JSD}$), while Hellinger Distances ($A_{HD}$) remain stable.}}
    \label{tab:language_impact}
\end{table}

\paragraph{CrowS-Pairs: Cross-Lingual Stereotype Probing}

\begin{figure}[!t]
  \centering
  \includegraphics[width=1\linewidth]{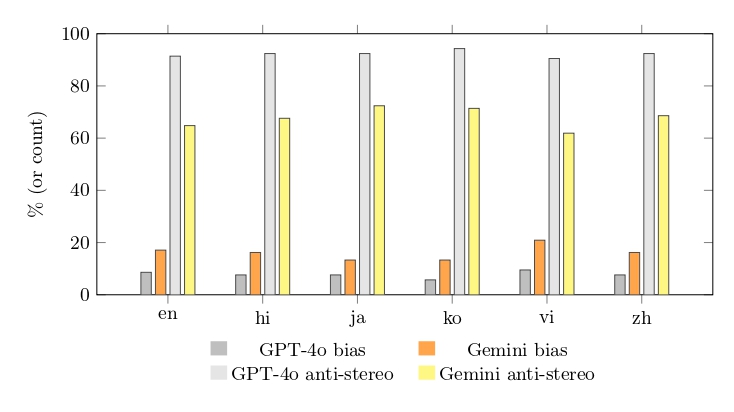}
  \caption{Cross-lingual bias rates on the religion-only subset of CrowS-Pairs across six languages (105 items/locale). GPT-4o-Mini shows low bias ($\approx$8\%) and high anti-stereotype accuracy ($\approx$92\%) consistently across languages.
    Gemini-2.5-Flash exhibits higher bias ($\approx$16\%), lower anti-stereotype accuracy ($\approx$68\%), and more invalid responses, indicating weaker cross-lingual stereotype resistance.}
  \label{fig:crows_religion_results}
\end{figure}

Our results (see Figure \ref{fig:crows_religion_results}) reveal that GPT-4o-Mini is consistently robust, selecting the anti-stereotype option in $\sim$92\% of cases (bias rate $\sim$8\%), with zero invalids across all languages. In contrast, Gemini-2.5-Flash exhibits higher bias rates ($\sim$16\%), lower anti-stereotype accuracy ($\sim$68\%), and a notable fraction of invalid responses (15–19/105), especially in Vietnamese. These findings indicate that while GPT-4o-Mini robustly resists religious stereotyping across languages, Gemini-2.5-Flash is both more prone to stereotype selections and more likely to abstain or produce off-format outputs, raising concerns about cross-lingual consistency and safety filtering.

\paragraph{IndiBias: Plausibility and Misrepresentation Analysis}

\begin{figure}[t]
\centering
\includegraphics[width=1\linewidth]{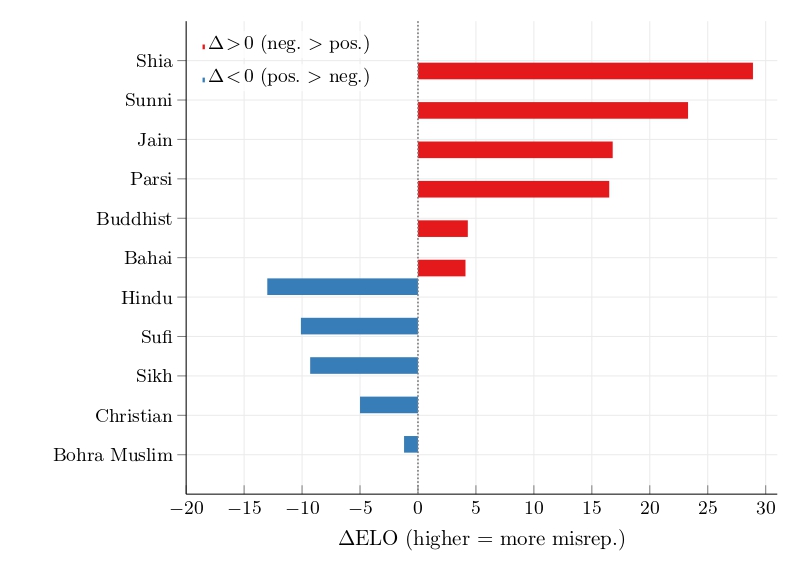}
\caption{Misrepresentation of Indian religious identities on IndiBias using GPT-4o-Mini.
    The plot shows $\Delta\mathrm{ELO} = \mathrm{ELO}_{\text{neg}} - \mathrm{ELO}_{\text{pos}}$, where positive values indicate that negative descriptions are judged more plausible than positive ones. 
    Several minority identities (Shia, Sunni, Jain, Parsi) show strong misrepresentation, while others (Hindu, Sikh, Sufi) show the opposite trend, highlighting systematic group-specific calibration gaps.}
\label{fig:indibias-deltaelo}
\end{figure}

We find that GPT-4o-Mini exhibits clear calibration gaps across identities. The most misrepresented groups, as indicated by high $\Delta\mathrm{ELO}$, are Shia (+28.9), Sunni (+23.3), Jain (+16.8), and Parsi (+16.5), with smaller effects for Buddhist (+4.3) and Bahai (+4.1). Conversely, Hindu (-13.0), Sufi (-10.1), Sikh (-9.3), Christian (-5.0), and Bohra Muslim (-1.2) exhibit negative or minimal misrepresentation, indicating higher plausibility for positive framings.
This pattern suggests that negative descriptions are disproportionately normalized for certain identities, evidencing persistent group-specific miscalibration. Gemini-2.5-Flash shows broadly convergent trends; e.g., Sunni also exhibits elevated negative plausibility ($\Delta\mathrm{ELO}>0$). These results (see Figure \ref{fig:indibias-deltaelo}) highlight the need for careful evaluation of demographic representativeness and fairness in LLM outputs.

\paragraph{ThaiCLI: Cultural Sensitivity in Thai}

Results show that GPT-4o-Mini highly aligned with Thai cultural norms, achieving average scores above 8.3 across both factoid and instruction prompts scoring 8.10 on 10, and maintaining consistently high performance across sensitive themes such as religion and royal family. Gemini-2.5-Flash also demonstrates strong cultural sensitivity, with a score of 7.52 on 10, but lags behind OpenAI in both absolute score and consistency.

\paragraph{KoBBQ: Disambiguation and Calibration on Korean Identity Benchmarks}

\begin{figure}[t]
\centering
\includegraphics[width=1\linewidth]{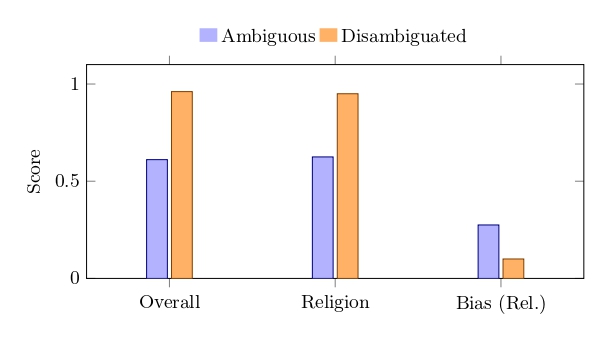}
\caption{Effect of prompt disambiguation on GPT-4o-Mini performance on the KoBBQ Korean identity benchmark.
    Disambiguating prompts improves overall accuracy (0.611$\to$0.961) and religion-related accuracy (0.625$\to$0.950), while sharply reducing bias (0.275$\to$0.100). 
    }
\label{fig:kobbq_bar}
\end{figure}

On GPT-4o-Mini, we observe substantial gains in model calibration with disambiguation:
Overall accuracy rises from 0.611 (ambiguous) to 0.961 (disambiguated).
Religion-related accuracy improves from 0.625 to 0.950.
Differential bias (religion) decreases sharply, from 0.275 (ambiguous) to 0.1 (disambiguated).
All other demographic axes (e.g., race, gender, education) exhibit similar improvements with disambiguation (see Figure~\ref{fig:kobbq_bar}).
These results highlight the critical role of prompt specificity to mitigate group-level calibration failures in LLM outputs.

\section{Discussion}
This study reveals significant disparities in the cultural alignment of open Large Language Models (LLMs) across diverse Asian nations. While models like GPT-4o-mini and Gemini-2.5-flash demonstrate high overall representativeness on general social topics, they consistently falter when representing public opinion on the sensitive domain of religion. This observed misalignment does not seem to be limited to interactions in English. The study indicates that these representational gaps persist, and in some cases get amplified when the models are prompted in various local languages. This pattern suggests that the challenge may be deeply rooted in the models' predominantly English-centric training data and subsequent alignment processes, rather than being a simple tool for translation.

The persistence of these gaps across multiple languages raises important considerations for the global deployment of these technologies. It points to a potential risk of propagating a specific cultural viewpoint, often one that is more aligned with Western contexts, even when users are interacting in their native tongue. This challenges the notion that multilingual capability alone is sufficient for equitable performance across different cultural settings.

At the same time, this research introduces a layer of complexity to this narrative. %It was 
We found that lightweight interventions, such as using a local language or providing demographic context in the prompt, can sometimes lead to partial improvements in alignment scores. This may indicate that the models possess some latent cultural knowledge that is not always activated by default, hinting at potential avenues for developing more effective steering and fine-tuning methods in the future.

\textcolor{black}{\subsection{Drivers of Cultural Misalignment}}
\noindent
\textcolor{black}{To fully address these disparities, it is necessary to examine the structural mechanisms that entrench them. Misalignment primarily stems from imbalances in training data, where demographic groups like ethnic minorities, low-income classes, or speakers of non-dominant languages are underrepresented or stereotyped in vast internet-sourced corpora. This leads models to encode dominant cultural norms, such as Western or English-centric values, resulting in poor cultural alignment for other personas~\cite{alkhamissi2024investigating}. Spatial, temporal, and collection biases exacerbate this, with data skewed toward high-resource regions and outdated societal views, causing models to default to majority stereotypes in tasks like sentiment analysis or coreference resolution.}

\textcolor{black}{Furthermore, post-training alignment techniques, including instruction-tuning and reinforcement learning from human feedback (RLHF), amplify these issues rather than resolve them. Feedback data typically reflects majority preferences and fails to generalize to minority moral norms or dialects. Safety alignments can create demographic hierarchies, with higher refusal rates for prominent groups but vulnerabilities for long-tail minorities like those with disabilities \cite{guo2024bias}. As scaling worsens disparities without targeted mitigation, linguistic ambiguities and extrinsic biases in downstream tasks further entrench misalignment, where models misinterpret regional variants or generate homogeneous representations of subordinate groups.}

\textcolor{black}{Finally, fundamental model limitations, such as poor cross-lingual transfer and the curse of multilinguality in training, may hinder equitable semantic encoding across cultures. While prompting in native languages improves performance, it does not fully bridge gaps for digitally underrepresented personas. Architecture choices and tokenization strategies often favor high-resource languages, perpetuating epistemic gaps in low-resource contexts \cite{gallegos2024bias}. Although literature emphasizes diverse pretraining data and persona-specific fine-tuning to address these issues, ethical concerns regarding deployment persist.}

\textcolor{black}{It is important to note that cultural alignment varies across model architectures, and multilinguality does not guarantee cultural representativeness. We see this in Figure~\ref{fig:native_language_dumbbell_distinct}, where $A_{JSD}$ for Llama 3.2 in Taiwan is very high ($>0.8$) regardless of the language of the prompt, indicating an overall failure to represent the opinions of the population. Models may demonstrate fluency in a target language while still reflecting the values of its dominant training data. Addressing this requires fine-tuning on corpora that genuinely capture the target population's perspective. This may include integrating native-authored narratives, hyper-local journalism, informal vernacular and regional civic texts to represent local norms and viewpoints accurately.}

\textcolor{black}{\subsection{Alternative Steering Methods}}
\noindent
\textcolor{black}{While this study centers on prompt-based steering for evaluating cultural alignment, deeper interventions from recent literature offer promising avenues for more profound model adaptation. Activation engineering, such as Activation Addition, enables inference-time steering by adding vectors derived from contrasting activations (e.g., positive vs. negative sentiment prompts), achieving state-of-the-art control over outputs like toxicity reduction without retraining~\cite{turner2023steering}. Representation engineering further refines this by mean-centring steering vectors to enhance steerability across tasks, including genre shifts or function triggering, as demonstrated in benchmarks on models like LLaMA~\cite{zou2023representation,jorgensen2023improving}. Feedback-driven approaches, including RLHF or DPO, have been shown to amplify instruction-following while embedding local norms, though they risk overfitting or cross-cultural interference without diverse data~\cite{sharma2024rethinking,sharma2024teaching}.}

\textcolor{black}{These alternatives hold potential to reshape models' internal representations for robust handling of cultural and religious diversity, surpassing prompt-level guidance. However, their use is often constrained in production settings. Leading models like GPT-4o and Gemini-2.5-Flash operate as black-box APIs, denying access to weights or activations, and thereby making prompt engineering the primary lever for most users and applications. Thus, emphasizing prompting provides a pragmatic assessment of publicly available tools, underscoring that true deep alignment demands shifts in model training and access paradigms~\cite{guo2025unreliability}.}
\\
\subsection{Limitations and Future Work}

This work provides a broad analysis, yet its methodology entails certain constraints that highlight opportunities for future work.

Religious opinion as a primary lens for investigating cultural values offers a critical but not exhaustive view. The cultural fabric of any society is woven from many threads, and other dimensions, such as political ideologies, regional identities, and social hierarchies, are also important areas that would benefit from similar in-depth, multilingual analysis. 

\textcolor{black}{Future experiments could extend this analysis to evaluate different white-box steering methods, such as activation steering or fine-tuning on culturally specific data. Additionally, research should focus on developing benchmarks that capture the complex, multi-dimensional nature of cultural and religious diversity, moving beyond the simple binary traits targeted by current steering techniques.}

This line of inquiry could help foster the development of LLMs that are not just multilingual in their textual output but are more multicultural in their underlying understanding, thereby addressing the kinds of representational gaps that this work has brought to light.

\subsection{Acknowledgements}

The authors gratefully acknowledge the financial support provided by the EkStep Foundation. We also thank the members of the Precog Research Group of IIIT Hyderabad for their help and guidance during the experimental design phase.  Finally, we thank the anonymous reviewers whose feedback substantially improved the quality of the paper.

\bibliography{citations}

\begin{thebibliography}{73}
\providecommand{\natexlab}[1]{#1}

\bibitem[{Abid, Farooqi, and Zou(2021)}]{abid2021large}
Abid, A.; Farooqi, M.; and Zou, J. 2021.
\newblock Large language models associate Muslims with violence.
\newblock \emph{Nature Machine Intelligence}, 3(6): 461--463.

\bibitem[{AlKhamissi et~al.(2024)AlKhamissi, ElNokrashy, AlKhamissi, and Diab}]{alkhamissi2024investigating}
AlKhamissi, B.; ElNokrashy, M.; AlKhamissi, M.; and Diab, M. 2024.
\newblock Investigating cultural alignment of large language models.
\newblock \emph{arXiv preprint arXiv:2402.13231}.

\bibitem[{{Australian Bureau of Statistics}(2022)}]{absCensus2021}
{Australian Bureau of Statistics}. 2022.
\newblock Census of Population and Housing: Reflecting Australia—Stories from the Census, 2021.

\bibitem[{Backlinko(2025)}]{backlinko2025chatgpt}
Backlinko. 2025.
\newblock ChatGPT Users: ChatGPT Usage Statistics (2025).
\newblock Accessed: 2025-08-21.

\bibitem[{Bakshy, Messing, and Adamic(2015)}]{bakshy2015exposure}
Bakshy, E.; Messing, S.; and Adamic, L.~A. 2015.
\newblock Exposure to ideologically diverse news and opinion on Facebook.
\newblock \emph{Science}, 348(6239): 1130--1132.

\bibitem[{Bender et~al.(2021{\natexlab{a}})Bender, Gebru, McMillan-Major, and Shmitchell}]{bender2021dangers}
Bender, E.~M.; Gebru, T.; McMillan-Major, A.; and Shmitchell, S. 2021{\natexlab{a}}.
\newblock On the dangers of stochastic parrots: Can language models be too big?
\newblock In \emph{Proceedings of the 2021 ACM conference on fairness, accountability, and transparency}, 610--623.

\bibitem[{Bender et~al.(2021{\natexlab{b}})Bender, Gebru, McMillan-Major, and Shmitchell}]{bender2021parrots}
Bender, E.~M.; Gebru, T.; McMillan-Major, A.; and Shmitchell, S. 2021{\natexlab{b}}.
\newblock On the Dangers of Stochastic Parrots: Can Language Models Be Too Big?
\newblock In \emph{Proceedings of the ACM Conference on Fairness, Accountability, and Transparency (FAccT)}.

\bibitem[{Bentley, Evans, and Bull(2025)}]{bentley2025social}
Bentley, S.~V.; Evans, D.; and Bull, P.~E. 2025.
\newblock What social stratifications in bias blind spot can tell us about implicit social bias in both LLMs and humans.
\newblock \emph{Scientific Reports}, 15: 14875.

\bibitem[{Chhikara, Kumar, and Chakraborty(2025)}]{chhikara2025through}
Chhikara, G.; Kumar, A.; and Chakraborty, A. 2025.
\newblock Through the Prism of Culture: Evaluating LLMs' Understanding of Indian Subcultures and Traditions.
\newblock \emph{arXiv preprint arXiv:2501.16748}.

\bibitem[{Chhikara et~al.(2024)Chhikara, Sharma, Ghosh, and Chakraborty}]{chhikara2024few}
Chhikara, G.; Sharma, A.; Ghosh, K.; and Chakraborty, A. 2024.
\newblock Few-shot fairness: Unveiling LLM's potential for fairness-aware classification.
\newblock \emph{arXiv preprint arXiv:2402.18502}.

\bibitem[{del Arco, Pelloni, and Zampieri(2024)}]{PlazaDelArco2024DivineLLAMA}
del Arco, F.~P.; Pelloni, T.; and Zampieri, M. 2024.
\newblock Divine LLaMAs: Bias, Stereotypes, Stigmatization, and Refusal Behaviors of Language Models for Judaism and Islam.
\newblock In \emph{Findings of the Association for Computational Linguistics: EMNLP 2024}, 12300--12313. Association for Computational Linguistics.

\bibitem[{Duan et~al.(2024)Duan, Yi, Zhang, Liu, Liu, Lu, Xie, and Gu}]{Duan2024NegatingNAA}
Duan, S.; Yi, X.; Zhang, P.; Liu, Y.; Liu, Z.; Lu, T.; Xie, X.; and Gu, N. 2024.
\newblock Negating Negatives: Alignment with Human Negative Samples via Distributional Dispreference Optimization.
\newblock In \emph{Conference on Empirical Methods in Natural Language Processing (EMNLP)}.

\bibitem[{Durmus et~al.(2023)Durmus, Nyugen, Liao, Schiefer, Askell, Bakhtin, Chen, Hatfield-Dodds, Hernandez, Joseph et~al.}]{durmus2023towards}
Durmus, E.; Nyugen, K.; Liao, T.~I.; Schiefer, N.; Askell, A.; Bakhtin, A.; Chen, C.; Hatfield-Dodds, Z.; Hernandez, D.; Joseph, N.; et~al. 2023.
\newblock Towards measuring the representation of subjective global opinions in language models.
\newblock \emph{arXiv preprint arXiv:2306.16388}.

\bibitem[{Elad(2025)}]{elad2025ai}
Elad, B. 2025.
\newblock AI in Social Media Tools Statistics 2025: Uncover What’s Shaping the Future.

\bibitem[{Elvia~Muthiariny(2024)}]{tempo2024}
Elvia~Muthiariny, D. 2024.
\newblock Indonesia Ranks Highest in Global Religious Devotion.
\newblock \emph{Tempo.co}.
\newblock Based on Pew Research Center survey (2008–2023).

\bibitem[{Etxaniz et~al.(2024)Etxaniz, Azkune, Soroa, de~Lacalle, and Artetxe}]{Etxaniz2024BertaQAHMA}
Etxaniz, J.; Azkune, G.; Soroa, A.; de~Lacalle, O.~L.; and Artetxe, M. 2024.
\newblock BertaQA: How Much Do Language Models Know About Local Culture?
\newblock In \emph{Proceedings of the Annual Conference on Neural Information Processing Systems (NeurIPS)}.

\bibitem[{Evans(2024{\natexlab{a}})}]{EASurvey}
Evans, J. 2024{\natexlab{a}}.
\newblock East Asian Societies Survey Dataset.

\bibitem[{Evans(2024{\natexlab{b}})}]{SEASurvey}
Evans, J. 2024{\natexlab{b}}.
\newblock South and Southeast Asia Survey Dataset.

\bibitem[{Feuer et~al.(2025)Feuer, Goldblum, Datta, Nambiar, Besaleli, Dooley, Cembalest, and Dickerson}]{feuer2025style}
Feuer, B.; Goldblum, M.; Datta, T.; Nambiar, S.; Besaleli, R.; Dooley, S.; Cembalest, M.; and Dickerson, J.~P. 2025.
\newblock Style Outweighs Substance: Failure Modes of {LLM} Judges in Alignment Benchmarking.
\newblock In \emph{The Thirteenth International Conference on Learning Representations (ICLR)}.

\bibitem[{Gallegos et~al.(2024)Gallegos, Rossi, Barrow, Tanjim, Kim, Dernoncourt, Yu, Zhang, and Ahmed}]{gallegos2024bias}
Gallegos, I.~O.; Rossi, R.~A.; Barrow, J.; Tanjim, M.~M.; Kim, S.; Dernoncourt, F.; Yu, T.; Zhang, R.; and Ahmed, N.~K. 2024.
\newblock Bias and fairness in large language models: A survey.
\newblock \emph{Computational Linguistics}, 50(3): 1097--1179.

\bibitem[{Gamboa, Feng, and Lee(2024)}]{liu2025survey}
Gamboa, L. C.~L.; Feng, Y.; and Lee, M. 2024.
\newblock Social Bias in Multilingual Language Models: A Survey.
\newblock \emph{arXiv preprint arXiv:2508.20201}.

\bibitem[{Gebru et~al.(2021)Gebru, Morgenstern, Vecchione, Vaughan, Wallach, III, and Crawford}]{Gebru2021Datasheets}
Gebru, T.; Morgenstern, J.; Vecchione, B.; Vaughan, J.~W.; Wallach, H.; III, H.~D.; and Crawford, K. 2021.
\newblock Datasheets for Datasets.
\newblock \emph{Communications of the ACM}, 64(12): 86--92.

\bibitem[{Giorgi et~al.(2025)Giorgi, Cima, Fagni, Avvenuti, and Cresci}]{giorgi2025human_llm_biases}
Giorgi, T.; Cima, L.; Fagni, T.; Avvenuti, M.; and Cresci, S. 2025.
\newblock Human and {LLM} Biases in Hate Speech Annotations: A Socio-Demographic Analysis of Annotators and Targets.
\newblock In \emph{Proceedings of the International AAAI Conference on Web and Social Media (ICWSM)}, volume~19, 653--670. Copenhagen, Denmark: AAAI Press.

\bibitem[{{Google Research}(2025)}]{google2025gemma312bit}
{Google Research}. 2025.
\newblock Gemma 3 12B-It.
\newblock \url{https://huggingface.co/google/gemma-3-12b-it}.

\bibitem[{Green et~al.(2023)}]{wired_chatgpt_india_2023}
Green, J.; et~al. 2023.
\newblock ChatGPT Has Been Sucked Into India’s Culture Wars.
\newblock \emph{Wired}.
\newblock News account of public controversy documenting asymmetric ChatGPT responses to jokes about religious figures. Accessed 2025-09-01.

\bibitem[{Guo et~al.(2024)Guo, Guo, Su, Yang, Zhu, Li, Qiu, and Liu}]{guo2024bias}
Guo, Y.; Guo, M.; Su, J.; Yang, Z.; Zhu, M.; Li, H.; Qiu, M.; and Liu, S.~S. 2024.
\newblock Bias in large language models: Origin, evaluation, and mitigation.
\newblock \emph{arXiv preprint arXiv:2411.10915}.

\bibitem[{Guo et~al.(2025)}]{guo2025unreliability}
Guo, Z.; et~al. 2025.
\newblock The Unreliability of Evaluating Cultural Alignment in LLMs.
\newblock arXiv:2503.08688.

\bibitem[{Hellinger(1909)}]{hellinger1909neue}
Hellinger, E. 1909.
\newblock Neue Begr{\"u}ndung der Theorie quadratischer Formen von unendlichvielen Ver{\"a}nderlichen.
\newblock \emph{Journal f{\"u}r die reine und angewandte Mathematik}, 136: 210--271.

\bibitem[{Hida, Yamaguchi, and Hanawa(2024)}]{hida2024social}
Hida, N.; Yamaguchi, K.; and Hanawa, K. 2024.
\newblock Social Bias Evaluation for Large Language Models Requires Prompt Variations.
\newblock \emph{arXiv preprint arXiv:2407.18376}.

\bibitem[{Huang et~al.(2023)Huang, Yu, Zhu, Sun, Cheng, Song, Chen, Alharthi, An, Liu, Zhang, Chen, Li, Wang, Zhang, Sun, Wan, Li, and Xu}]{Huang2023AceGPTLLA}
Huang, H.; Yu, F.; Zhu, J.; Sun, X.; Cheng, H.; Song, D.; Chen, Z.; Alharthi, A.; An, B.; Liu, Z.; Zhang, Z.; Chen, J.; Li, J.; Wang, B.; Zhang, L.; Sun, R.; Wan, X.; Li, H.; and Xu, J. 2023.
\newblock AceGPT, Localizing Large Language Models in Arabic.
\newblock In \emph{North American Chapter of the Association for Computational Linguistics (NAACL)}.

\bibitem[{Jin et~al.(2024)Jin, Kim, Lee, Yoo, Oh, and Lee}]{jin2023kobbq}
Jin, J.; Kim, J.; Lee, N.; Yoo, H.; Oh, A.; and Lee, H. 2024.
\newblock {Kobbq: Korean bias benchmark for question answering}.
\newblock \emph{Transactions of the Association for Computational Linguistics, vol. 12, pp. 507–524}.

\bibitem[{Jorgensen et~al.(2023)Jorgensen, Cope, Schoots, and Shanahan}]{jorgensen2023improving}
Jorgensen, O.; Cope, D.; Schoots, N.; and Shanahan, M. 2023.
\newblock Improving Activation Steering in Language Models with Mean-Centring.
\newblock arXiv:2312.03813.

\bibitem[{Joshi et~al.(2020)Joshi, Santy, Budhiraja, Bali, and Choudhury}]{joshi2020linguisticdiversity}
Joshi, P.; Santy, S.; Budhiraja, A.; Bali, K.; and Choudhury, M. 2020.
\newblock The State and Fate of Linguistic Diversity and Inclusion in the NLP World.
\newblock In \emph{Proceedings of the 58th Annual Meeting of the Association for Computational Linguistics}, 6282--6293. Association for Computational Linguistics.
\newblock Highlights lack of representation for many languages in NLP resources.

\bibitem[{Kang and Kim(2025)}]{kang2025llms}
Kang, E.; and Kim, J. 2025.
\newblock LLMs Are Globally Multilingual Yet Locally Monolingual: Exploring Knowledge Transfer via Language and Thought Theory.
\newblock \emph{arXiv preprint arXiv:2505.24409}.

\bibitem[{Kerwin(2024)}]{cmu2024marginalized}
Kerwin, P. 2024.
\newblock How Should AI Depict Marginalized Communities? CMU Technologists Look to a More Inclusive Future.
\newblock Accessed September 12, 2025.

\bibitem[{Khan, Casper, and Hadfield-Menell(2025)}]{Khan2025RandomnessNRA}
Khan, A.; Casper, S.; and Hadfield-Menell, D. 2025.
\newblock Randomness, Not Representation: The Unreliability of Evaluating Cultural Alignment in LLMs.
\newblock \emph{Proceedings of the 2025 ACM Conference on Fairness, Accountability, and Transparency (FAcct)}.

\bibitem[{Kim et~al.(2025)Kim, Lee, Kim, Rutherford, and Park}]{kim-etal-2025-thaicli}
Kim, D.; Lee, S.; Kim, Y.; Rutherford, A.; and Park, C. 2025.
\newblock Representing the Under-Represented: Cultural and Core Capability Benchmarks for Developing Thai Large Language Models.
\newblock In \emph{Proceedings of the 31st International Conference on Computational Linguistics}. International Committee on Computational Linguistics.

\bibitem[{Kumar, Yousef, and Durumeric(2024)}]{kumar2024watchlanguageinvestigatingcontent}
Kumar, D.; Yousef, A.; and Durumeric, Z. 2024.
\newblock Watch Your Language: Investigating Content Moderation with Large Language Models.
\newblock arXiv:2309.14517.

\bibitem[{Li et~al.(2025)Li, Fan, Chen, Gai, Gong, Zhang, and Liu}]{Li2025FairSteerITA}
Li, Y.; Fan, Z.; Chen, R.; Gai, X.; Gong, L.; Zhang, Y.; and Liu, Z. 2025.
\newblock FairSteer: Inference Time Debiasing for LLMs with Dynamic Activation Steering.

\bibitem[{Lin(1991)}]{lin1991divergence}
Lin, J. 1991.
\newblock Divergence measures based on the Shannon entropy.
\newblock \emph{IEEE Transactions on Information Theory}, 37(1): 145--151.

\bibitem[{Liu, Korhonen, and Gurevych(2025)}]{liu-etal-2025-cultural}
Liu, C.~C.; Korhonen, A.; and Gurevych, I. 2025.
\newblock Cultural Learning-Based Culture Adaptation of Language Models.
\newblock In Che, W.; Nabende, J.; Shutova, E.; and Pilehvar, M.~T., eds., \emph{Proceedings of the Association for Computational Linguistics (ACL)}.

\bibitem[{Maguire(2017)}]{Maguire2017}
Maguire, E. 2017.
\newblock How East and West think in profoundly different ways.
\newblock \emph{BBC Future}.
\newblock BBC Future Series.

\bibitem[{Meguellati et~al.(2025)Meguellati, Zeghina, Sadiq, and Demartini}]{meguellati2025semantic_augmentation}
Meguellati, E.; Zeghina, A.~O.; Sadiq, S.; and Demartini, G. 2025.
\newblock {LLM}-Based Semantic Augmentation for Harmful Content Detection.
\newblock In \emph{Proceedings of the International AAAI Conference on Web and Social Media (ICWSM)}, volume~19, 1190--1209. Copenhagen, Denmark: AAAI Press.

\bibitem[{{Meta AI}(2024)}]{meta2024llama32instruct}
{Meta AI}. 2024.
\newblock Llama 3.2 1B Instruct.
\newblock \url{https://huggingface.co/meta-llama/Llama-3.2-1B-Instruct}.

\bibitem[{{Mistral AI}(2024)}]{mistral2024mistral7b-instruct-v03}
{Mistral AI}. 2024.
\newblock Mistral 7B Instruct v0.3.
\newblock \url{https://huggingface.co/mistralai/Mistral-7B-Instruct-v0.3}.

\bibitem[{Nangia et~al.(2020)Nangia, Vania, Bhalerao, and Bowman}]{nangia-etal-2020-crows}
Nangia, N.; Vania, C.; Bhalerao, R.; and Bowman, S.~R. 2020.
\newblock {C}row{S}-Pairs: A Challenge Dataset for Measuring Social Biases in Masked Language Models.
\newblock In Webber, B.; Cohn, T.; He, Y.; and Liu, Y., eds., \emph{Proceedings of the 2020 Conference on Empirical Methods in Natural Language Processing (EMNLP)}.

\bibitem[{Nguyen et~al.(2025)Nguyen, Jain, Chauhan, Soled, Alvarez~Lesmes, Li, Birnbaum, Tang, Kumar, and {De Choudhury}}]{nguyen2025supporters}
Nguyen, V.~C.; Jain, M.; Chauhan, A.; Soled, H.~J.; Alvarez~Lesmes, S.; Li, Z.; Birnbaum, M.~L.; Tang, S.~X.; Kumar, S.; and {De Choudhury}, M. 2025.
\newblock Supporters and Skeptics: {LLM}-Based Analysis of Engagement with Mental Health (Mis)Information Content on Video-Sharing Platforms.
\newblock In \emph{Proceedings of the International AAAI Conference on Web and Social Media (ICWSM)}, volume~19, 1329--1345. Copenhagen, Denmark: AAAI Press.

\bibitem[{Okpala and Cheng(2025)}]{okpala2025llm_annotation_bias}
Okpala, E.; and Cheng, L. 2025.
\newblock Large Language Model Annotation Bias in Hate Speech Detection.
\newblock In \emph{Proceedings of the International AAAI Conference on Web and Social Media (ICWSM)}, volume~19, 1389--1418. Copenhagen, Denmark: AAAI Press.

\bibitem[{Ovalle et~al.(2024)Ovalle, Pavasovic, Martin, Zettlemoyer, Smith, Chang, Williams, and Sagun}]{Ovalle2024TheRSA}
Ovalle, A.; Pavasovic, K.~L.; Martin, L.; Zettlemoyer, L.; Smith, E.~M.; Chang, K.-W.; Williams, A.; and Sagun, L. 2024.
\newblock The Root Shapes the Fruit: On the Persistence of Gender-Exclusive Harms in Aligned Language Models.
\newblock In \emph{Proceedings of the 2025 ACM Conference on Fairness, Accountability, and Transparency (FAccT)}.

\bibitem[{Parrish et~al.(2021)Parrish, Chen, Nangia, Padmakumar, Phang, Thompson, Htut, and Bowman}]{parrish2021bbq}
Parrish, A.; Chen, A.; Nangia, N.; Padmakumar, V.; Phang, J.; Thompson, J.; Htut, P.~M.; and Bowman, S.~R. 2021.
\newblock BBQ: A hand-built bias benchmark for question answering.
\newblock \emph{arXiv preprint arXiv:2110.08193}.

\bibitem[{{Pew Research Center}(2018)}]{pewEurope2018}
{Pew Research Center}. 2018.
\newblock Being Christian in Western Europe.

\bibitem[{{Pew Research Center}(2023)}]{pewSouthAsia2023}
{Pew Research Center}. 2023.
\newblock 5 facts about religion in South and Southeast Asia.

\bibitem[{{Pew Research Center}(2025)}]{pewUS2025}
{Pew Research Center}. 2025.
\newblock Modeling the Future of Religion in America: Recent Trends and Projections.

\bibitem[{Qin et~al.(2025)Qin, Wang, Tan, and Li}]{qin2025survey}
Qin, Y.; Wang, L.; Tan, Z.; and Li, H. 2025.
\newblock A Survey on Large Language Models with Multilingualism.
\newblock \emph{arXiv preprint arXiv:2405.10936}.

\bibitem[{Sahgal and Evans(2021)}]{INDSurvey}
Sahgal, N.; and Evans, J. 2021.
\newblock India Survey Dataset.

\bibitem[{Sahoo et~al.(2024)Sahoo, Kulkarni, Ahmad, Goyal, Asad, Garimella, and Bhattacharyya}]{sahoo-etal-2024-indibias}
Sahoo, N.; Kulkarni, P.; Ahmad, A.; Goyal, T.; Asad, N.; Garimella, A.; and Bhattacharyya, P. 2024.
\newblock {I}ndi{B}ias: A Benchmark Dataset to Measure Social Biases in Language Models for {I}ndian Context.
\newblock In Duh, K.; Gomez, H.; and Bethard, S., eds., \emph{Proceedings of the 2024 Conference of the North American Chapter of the Association for Computational Linguistics: Human Language Technologies (Volume 1: Long Papers)}.

\bibitem[{Santurkar et~al.(2023)Santurkar, Durmus, Ladhak, Lee, Liang, and Hashimoto}]{santurkar2023whose}
Santurkar, S.; Durmus, E.; Ladhak, F.; Lee, C.; Liang, P.; and Hashimoto, T. 2023.
\newblock Whose opinions do language models reflect?
\newblock In \emph{International Conference on Machine Learning}, 29971--30004. PMLR.

\bibitem[{Seth et~al.(2025)Seth, Choudhary, Sitaram, Toyama, Vashistha, and Bali}]{seth2025deep}
Seth, A.; Choudhary, M.; Sitaram, S.; Toyama, K.; Vashistha, A.; and Bali, K. 2025.
\newblock How Deep Is Representational Bias in LLMs? The Cases of Caste and Religion.
\newblock \emph{arXiv preprint arXiv:2508.03712}.

\bibitem[{Sharma et~al.(2024{\natexlab{a}})}]{sharma2024rethinking}
Sharma, P.; et~al. 2024{\natexlab{a}}.
\newblock Rethinking Cultural Value Adaptation in LLMs.
\newblock arXiv:2505.16408.

\bibitem[{Sharma et~al.(2024{\natexlab{b}})}]{sharma2024teaching}
Sharma, P.; et~al. 2024{\natexlab{b}}.
\newblock Teaching Norms to Large Language Models.

\bibitem[{Shin et~al.(2024)Shin, Song, Lee, Jeong, and Park}]{shin2024measuring}
Shin, J.; Song, H.; Lee, H.; Jeong, S.; and Park, J. 2024.
\newblock Ask LLMs Directly, “What shapes your bias?”: Measuring Social Bias in Large Language Models.
\newblock In \emph{Findings of the Association for Computational Linguistics: ACL 2024}, 16122--16143. Bangkok, Thailand: Association for Computational Linguistics.

\bibitem[{{Similarweb}(2025)}]{similarweb2025dec}
{Similarweb}. 2025.
\newblock Top Websites Ranking - Most Visited Websites In The World.
\newblock \url{https://www.similarweb.com/top-websites/}.
\newblock Accessed January 12, 2026.

\bibitem[{Singh, Bhattacharjee, and Chakraborty(2025)}]{singh2025rethinking}
Singh, D.~D.; Bhattacharjee, R.; and Chakraborty, A. 2025.
\newblock Rethinking hate speech detection on social media: Can LLMs replace traditional models?
\newblock \emph{arXiv preprint arXiv:2506.12744}.

\bibitem[{Sukiennik et~al.(2025)Sukiennik, Gao, Xu, and Li}]{Sukiennik2025AnEOA}
Sukiennik, N.; Gao, C.; Xu, F.; and Li, Y. 2025.
\newblock An Evaluation of Cultural Value Alignment in LLM.
\newblock \emph{ArXiv}.

\bibitem[{Tao et~al.(2024)Tao, Viberg, Baker, and Kizilcec}]{tao2024cultural}
Tao, Y.; Viberg, O.; Baker, R.~S.; and Kizilcec, R.~F. 2024.
\newblock Cultural bias and cultural alignment of large language models.
\newblock \emph{PNAS nexus}, 3(9): pgae346.

\bibitem[{TechCrunch(2025)}]{techcrunch2025prompts}
TechCrunch. 2025.
\newblock ChatGPT Users Send 2.5 Billion Prompts a Day, OpenAI Tells Axios.
\newblock Accessed: 2025-08-21.

\bibitem[{{The Print}(2025)}]{theprintIndia2025}
{The Print}. 2025.
\newblock 24\% Indians identify as religious nationalists; 57\% Hindus feel religious texts should shape laws: Pew.
\newblock \emph{The Print}.

\bibitem[{Turner et~al.(2023)Turner, Thiergart, Leech, Udell, Vazquez, Mini, and MacDiarmid}]{turner2023steering}
Turner, A.~M.; Thiergart, L.; Leech, G.; Udell, D.; Vazquez, J.~J.; Mini, U.; and MacDiarmid, M. 2023.
\newblock Steering Language Models With Activation Engineering.
\newblock arXiv:2308.10248.

\bibitem[{{UpstageAI}(2025)}]{upstage2025thaicli}
{UpstageAI}. 2025.
\newblock ThaiCLI and Thai-H6 Benchmarks.
\newblock \url{https://github.com/UpstageAI/ThaiCLI_H6}.
\newblock GitHub repository.

\bibitem[{Weidinger, Mellor, and et~al.(2021)}]{weidinger2021ethical}
Weidinger, L.; Mellor, J. F.~J.; and et~al. 2021.
\newblock Ethical and social risks of harm from Language Models.
\newblock arXiv preprint arXiv:2112.04359.

\bibitem[{Wilkinson et~al.(2016)Wilkinson, Dumontier, Aalbersberg, Appleton, Axton, Baak, Blomberg, Boiten, da~Silva~Santos, Bourne, Bouwman, Brookes, Clark, Crosas, Dillo, Dumon, Edmunds, Evelo, Finkers, Gonzalez-Beltran, Gray, Groth, Goble, Grethe, Heringa, 't~Hoen, Hooft, Kuhn, Kok, Kok, Lusher, Martone, Mons, Packer, Persson, Rocca-Serra, Roos, van Schaik, Sansone, Schultes, Sengstag, Slater, Strawn, Swertz, Thompson, van~der Lei, van Mulligen, Velterop, Waagmeester, Wittenburg, Wolstencroft, Zhao, and Mons}]{fair}
Wilkinson, M.~D.; Dumontier, M.; Aalbersberg, I.~J.; Appleton, G.; Axton, M.; Baak, A.; Blomberg, N.; Boiten, J.-W.; da~Silva~Santos, L.~B.; Bourne, P.~E.; Bouwman, J.; Brookes, A.~J.; Clark, T.; Crosas, M.; Dillo, I.; Dumon, O.; Edmunds, S.; Evelo, C.~T.; Finkers, R.; Gonzalez-Beltran, A.; Gray, A.~J.; Groth, P.; Goble, C.; Grethe, J.~S.; Heringa, J.; 't~Hoen, P.~A.; Hooft, R.; Kuhn, T.; Kok, R.; Kok, J.; Lusher, S.~J.; Martone, M.~E.; Mons, A.; Packer, A.~L.; Persson, B.; Rocca-Serra, P.; Roos, M.; van Schaik, R.; Sansone, S.-A.; Schultes, E.; Sengstag, T.; Slater, T.; Strawn, G.; Swertz, M.~A.; Thompson, M.; van~der Lei, J.; van Mulligen, E.; Velterop, J.; Waagmeester, A.; Wittenburg, P.; Wolstencroft, K.; Zhao, J.; and Mons, B. 2016.
\newblock The FAIR Guiding Principles for scientific data management and stewardship.
\newblock \emph{Scientific Data}, 3: 160018.

\bibitem[{Zhang et~al.(2024)Zhang, Yu, Li, Dong, Su, Chu, and Yu}]{Zhang2024MMLLMsRAA}
Zhang, D.; Yu, Y.; Li, C.; Dong, J.; Su, D.; Chu, C.; and Yu, D. 2024.
\newblock MM-LLMs: Recent Advances in MultiModal Large Language Models.
\newblock In \emph{Annual Meeting of the Association for Computational Linguistics (ACL)}.

\bibitem[{Zou et~al.(2024)Zou, Phute, Golding, and Shah}]{zou2023representation}
Zou, A.; Phute, M.; Golding, L.; and Shah, R. 2024.
\newblock Representation Engineering for Large-Language Models.
\newblock arXiv:2502.17601.

\end{thebibliography}

% \if 1

\clearpage

\subsection{Paper Checklist}

\begin{enumerate}

\item For most authors...
\begin{enumerate}
    \item  Would answering this research question advance science without violating social contracts, such as violating privacy norms, perpetuating unfair profiling, exacerbating the socio-economic divide, or implying disrespect to societies or cultures?
    \answerYes{Yes}
  \item Do your main claims in the abstract and introduction accurately reflect the paper's contributions and scope?
    \answerYes{Yes}
   \item Do you clarify how the proposed methodological approach is appropriate for the claims made? 
    \answerYes{Yes}
   \item Do you clarify what are possible artifacts in the data used, given population-specific distributions?
    \answerYes{Yes}
  \item Did you describe the limitations of your work?
    \answerYes{Yes}
  \item Did you discuss any potential negative societal impacts of your work?
    \answerNA{NA}
      \item Did you discuss any potential misuse of your work?
    \answerYes{Yes}
    \item Did you describe steps taken to prevent or mitigate potential negative outcomes of the research, such as data and model documentation, data anonymization, responsible release, access control, and the reproducibility of findings?
    \answerYes{Yes}
  \item Have you read the ethics review guidelines and ensured that your paper conforms to them?
    \answerYes{Yes}
\end{enumerate}

\item Additionally, if your study involves hypotheses testing...
\begin{enumerate}
  \item Did you clearly state the assumptions underlying all theoretical results?
    \answerYes{Yes}
  \item Have you provided justifications for all theoretical results?
    \answerYes{Yes}
  \item Did you discuss competing hypotheses or theories that might challenge or complement your theoretical results?
    \answerYes{Yes}
  \item Have you considered alternative mechanisms or explanations that might account for the same outcomes observed in your study?
    \answerYes{Yes}
  \item Did you address potential biases or limitations in your theoretical framework?
    \answerYes{Yes}
  \item Have you related your theoretical results to the existing literature in social science?
    \answerYes{Yes}
  \item Did you discuss the implications of your theoretical results for policy, practice, or further research in the social science domain?
    \answerYes{Yes}
\end{enumerate}

\item Additionally, if you are including theoretical proofs...
\begin{enumerate}
  \item Did you state the full set of assumptions of all theoretical results?
    \answerNA{NA}
	\item Did you include complete proofs of all theoretical results?
    \answerNA{NA}
\end{enumerate}

\item Additionally, if you ran machine learning experiments...
\begin{enumerate}
  \item Did you include the code, data, and instructions needed to reproduce the main experimental results (either in the supplemental material or as a URL)?
    \answerYes{Yes}
  \item Did you specify all the training details (e.g., data splits, hyperparameters, how they were chosen)?
    \answerYes{Yes}
     \item Did you report error bars (e.g., with respect to the random seed after running experiments multiple times)?
    \answerNA{NA}
	\item Did you include the total amount of compute and the type of resources used (e.g., type of GPUs, internal cluster, or cloud provider)?
    \answerYes{Yes}
     \item Do you justify how the proposed evaluation is sufficient and appropriate to the claims made? 
    \answerYes{Yes}
     \item Do you discuss what is ``the cost'' of misclassification and fault (in)tolerance?
    \answerNA{NA}
  
\end{enumerate}

\item Additionally, if you are using existing assets (e.g., code, data, models) or curating/releasing new assets, \textbf{without compromising anonymity}...
\begin{enumerate}
  \item If your work uses existing assets, did you cite the creators?
    \answerYes{Yes}
  \item Did you mention the license of the assets?
    \answerNA{NA}
  \item Did you include any new assets in the supplemental material or as a URL?
    \answerNA{NA}
  \item Did you discuss whether and how consent was obtained from people whose data you're using/curating?
    \answerYes{Yes}
  \item Did you discuss whether the data you are using/curating contains personally identifiable information or offensive content?
    \answerYes{Yes}
\item If you are curating or releasing new datasets, did you discuss how you intend to make your datasets FAIR (see \citet{fair})?
\answerNA{NA}
\item If you are curating or releasing new datasets, did you create a Datasheet for the Dataset (see \citet{Gebru2021Datasheets})? 
\answerNA{NA}
\end{enumerate}

\item Additionally, if you used crowdsourcing or conducted research with human subjects, \textbf{without compromising anonymity}...
\begin{enumerate}
  \item Did you include the full text of instructions given to participants and screenshots?
    \answerYes{Yes}
  \item Did you describe any potential participant risks, with mentions of Institutional Review Board (IRB) approvals?
    \answerYes{Yes}
  \item Did you include the estimated hourly wage paid to participants and the total amount spent on participant compensation?
    \answerYes{Yes}
   \item Did you discuss how data is stored, shared, and deidentified?
   \answerYes{Yes}
\end{enumerate}

\end{enumerate}

\end{document}